%% file: main.tex
\documentclass[10pt,twocolumn,letterpaper]{article}

\usepackage{iccv}
\usepackage{times}
\usepackage{epsfig}
\usepackage{amsmath}
\usepackage{amssymb}
\usepackage{amsfonts}
\usepackage{bm}

\usepackage{authblk}
\usepackage{algorithm}
\usepackage{algorithmic}

\usepackage[numbers]{natbib}
\usepackage{multirow}
\usepackage{graphicx}
\usepackage{color}
\usepackage[it,small]{caption}
\usepackage{subcaption}
\usepackage{dsfont}
\usepackage{enumitem}
\usepackage{authblk}
\setlist{nolistsep}

\usepackage[breaklinks=true,bookmarks=false]{hyperref}

\newcommand{\argmax}{\operatorname{arg\,max}}

\input{space_saver}
\newcommand{\todo}[1]{\textcolor{blue}{\textbf{#1}}}

\iccvfinalcopy 

\def\iccvPaperID{579} 

\ificcvfinal\fi
\begin{document}
\setlength{\abovedisplayskip}{0pt}
\setlength{\belowdisplayskip}{0pt}
\setlength{\abovedisplayshortskip}{0pt}
\setlength{\belowdisplayshortskip}{0pt}

\title{Car that Knows Before You Do: \\Anticipating Maneuvers via Learning Temporal Driving Models}
\author[1,2]{Ashesh Jain}
\author[1,2]{Hema S Koppula}
\author[1]{Bharad Raghavan}
\author[1]{Shane Soh}
\author[2,3]{Ashutosh Saxena}
\affil[ ]{Stanford University$^1$, Cornell University$^2$, Brain Of Things Inc.$^3$}
\affil[ ]{{\{ashesh,hema,bharadr,shanesoh,asaxena\}@cs.stanford.edu}}

\maketitle

\input{abstract}

\input{intro}

\input{relatedwork}

\input{overview}

\input{model}

\input{features}

\input{experiment}

\input{conclusions}

\end{document}

%% file: space_saver.tex
 \belowdisplayskip \abovedisplayskip
\renewcommand{\baselinestretch}{0.97}

 \newlength\savedwidth

\newlength{\sectionReduceTop}
\newlength{\sectionReduceBot}
\newlength{\subsectionReduceTop}
\newlength{\subsectionReduceBot}
\newlength{\abstractReduceTop}
\newlength{\abstractReduceBot}
\newlength{\captionReduceTop}
\newlength{\captionReduceBot}
\newlength{\subsubsectionReduceTop}
\newlength{\subsubsectionReduceBot}

\newlength{\horSkip}
\newlength{\verSkip}

\newlength{\figureHeight}

%% file: abstract.tex

\begin{abstract}
Advanced Driver Assistance Systems (ADAS) have made driving safer over the last decade. They prepare vehicles for  unsafe road conditions and alert drivers if they perform a dangerous maneuver.  However, many accidents are unavoidable because by the time drivers are alerted, it is already too late.  Anticipating maneuvers beforehand can alert drivers before they perform the  maneuver and also give ADAS more time to avoid or prepare for the danger.

In this work we anticipate driving maneuvers a few seconds before they occur. For this purpose we equip a car with cameras and a computing device to capture the driving context from both inside and outside of the car. We propose an Autoregressive Input-Output HMM to model the contextual information alongwith the maneuvers. We evaluate our approach on a  diverse data set with 1180 miles of natural freeway and city driving and show that we can anticipate  maneuvers 3.5 seconds before they occur with over 80\% F1-score in real-time. 

\end{abstract}

%% file: intro.tex
\vspace{\sectionReduceTop}
\section{Introduction}
\vspace{\sectionReduceBot}

\begin{figure}[t]
\centering
\vskip -0.05in
\includegraphics[width=.8\linewidth]{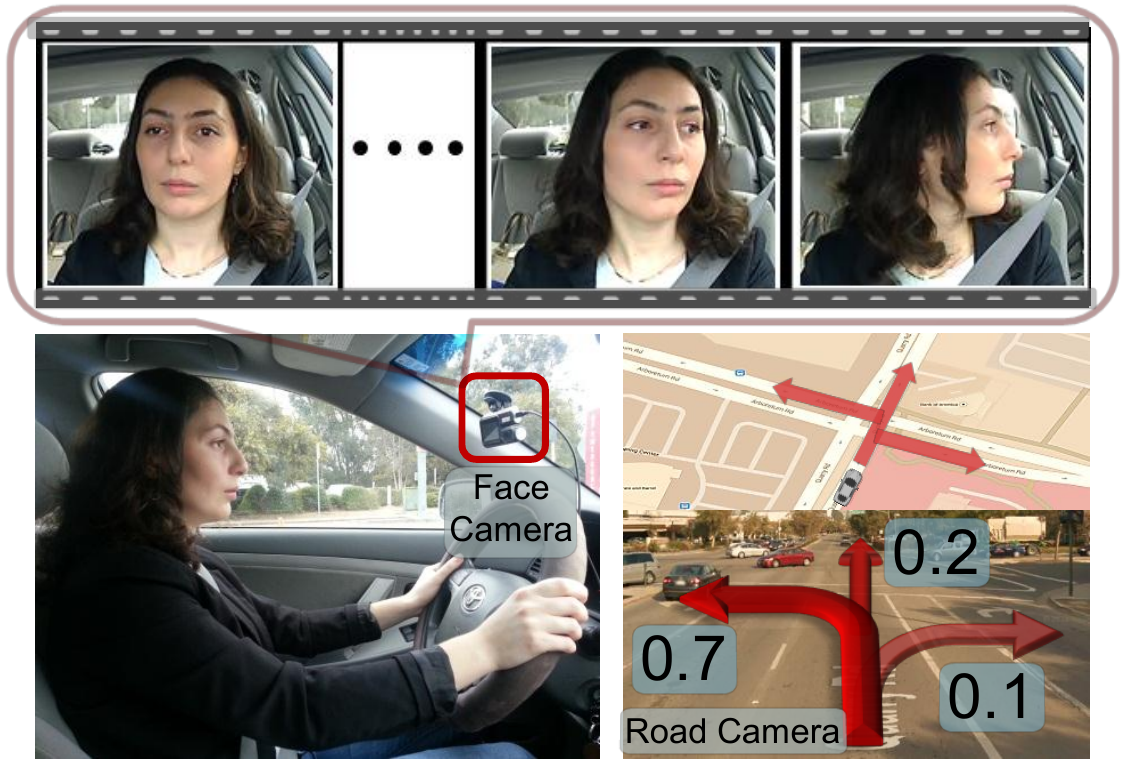}
\vspace{\captionReduceTop}
\caption{\textbf{Anticipating maneuvers.} Our algorithm anticipates driving maneuvers performed a few seconds in the future. It uses information from multiple sources including videos, vehicle dynamics, GPS, and street maps to anticipate the probability of different future maneuvers.}
\vspace{2\captionReduceBot}
\label{fig:intro}
\end{figure}

Over the last decade
cars have been equipped with various assistive technologies in order to
provide a safe driving experience. Technologies such as lane keeping, blind spot check,
pre-crash systems etc.,  are successful in alerting drivers whenever they commit a
dangerous maneuver~\citep{Laugier11}. Still in the US alone more than 33,000 people die in road
accidents every year, the majority of which are due to inappropriate
maneuvers~\cite{road_accidents}. 
We need mechanisms that	 can alert drivers \textit{before} they perform a dangerous maneuver 
in order to avert many such accidents~\citep{Rueda04}.  
In this work we address this problem of anticipating maneuvers that a
driver is likely to perform in the next few seconds (Figure~\ref{fig:intro}). 

Anticipating future human actions has recently been a topic of interest to both the vision and robotics 
communities~\citep{Kitani12,Koppula13,Wang13}. Figure~\ref{fig:intro} 
shows our system anticipating a left turn maneuver a few seconds before the car reaches the 
intersection. Our system also outputs probabilities over the  maneuvers the driver can perform. 
With this prior knowledge of maneuvers, the driver assistance systems can 
alert drivers about possible dangers before they perform the maneuver, 
thereby giving them more time to react. Some previous works~\citep{Frohlich14,Kumar13,Morris11} also predict a driver's  future maneuver.  However, as we show in the following sections, these methods use limited context and do not accurately model the anticipation problem.

In order to anticipate maneuvers, we reason with the contextual information from 
the surrounding events, which we refer to as the \textit{driving context}. 
We obtain this driving context from multiple sources. We use videos
of the driver inside the car and the road in front, the vehicle's dynamics,
global position coordinates (GPS), and street maps; from this we extract a
time series of multi-modal data from both inside and outside the vehicle. The
challenge lies in modeling the temporal aspects of driving and in detecting the contextual
cues that help in anticipating maneuvers. 

Modeling maneuver anticipation also requires joint reasoning of the driving
context and the driver's intention. The challenge here is the driver's
intentions are not directly observable, and their interactions with the driving
context are complex.
For example, the driver is influenced by external events such as traffic conditions.
The nature of these interactions is generative and they require a specially tailored modeling approach.

In this work we propose a model and a learning algorithm to  capture the temporal aspects of the problem, along with the generative nature of the interactions. Our model is an
Autoregressive Input-Output Hidden Markov Model (AIO-HMM) that jointly captures the context
from both inside and outside the vehicle.  AIO-HMM models how events from outside the vehicle affect the driver's intention, which then generates events inside the vehicle.
We learn the AIO-HMM model parameters from natural driving data and during inference output the probability of each maneuver.

We evaluate our approach on a driving data set with 1180 miles of natural freeway and
city driving collected across two states -- from 10 drivers and with different kinds of driving maneuvers.
We demonstrate that our approach anticipates 
maneuvers 3.5 seconds before they occur with 80\% precision and recall. We
believe that our work creates scope for new ADAS features to make roads safer.  
In summary our key contributions are as follows: 
\begin{itemize}
\itemsep0em 
\item We propose an approach for anticipating driving maneuvers several seconds in advance.
\item We model the driving context from inside and outside the car with an autoregressive input-output HMM. 
\item We release the first data set of natural driving with videos from both inside and outside the car, GPS, and speed information.
\end{itemize}
 Our data set and code are available at: \url{http://www.brain4cars.com}.

%% file: relatedwork.tex
\vspace{\sectionReduceTop}
\section{Related Work}
\vspace{\sectionReduceBot}
\noindent \textbf{Assistive features for vehicles.} 
Latest cars available in market comes equipped with cameras and sensors to monitor the surrounding environment. Through multi-sensory fusion they provide assisitive features like lane keeping, forward collision avoidance, adaptive cruise control etc. These systems warn drivers when they perform a potentially dangerous maneuver~\citep{Shia14,Vasudevan12}. Driver monitoring for distraction and drowsiness has also been extensively researched~\citep{Fletcher05,Rezaei14}. Techniques like eye-gaze tracking are now commercially available (Seeing Machines Ltd.) and has been effective in detecting distraction. Our work complements existing ADAS and driver monitoring techniques by anticipating maneuvers several seconds before they occur.

Closely related to us are previous works on predicting the driver's intent. Vehicle trajectory has been used to predict the intent for lane change or turn maneuver~\citep{Berndt08,Frohlich14,Kumar13,Liebner12}. Most of these works ignore the rich context available from cameras, GPS, and street maps. Trivedi et al.~\citep{Trivedi07} and Morris et al.~\citep{Morris11} predict lane change intent using the rich context from cameras both inside and outside the vehicle. Both works train a discriminative classifier which assumes that informative contextual cues always appear at a fixed time before the maneuver. We show that this assumption is not true, and in fact the temporal aspect of the problem should be carefully modeled. Our AIO-HMM takes a generative approach and handles the temporal aspect of this problem.

\noindent \textbf{Anticipation and Modeling Humans.} Modeling of human motion has given rise to many applications, anticipation being one of them. Wang et al.~\citep{Wang13}, Koppula et al.~\citep{Koppula13,Koppula15}, and Sener et al.~\citep{Sener15} 
demonstrate better human-robot collaboration by anticipating  a human's future movements. Kitani et al.~\citep{Kitani12} model human navigation in order to anticipate the path they will follow. Similar to these works, we anticipate human actions, which are driving maneuvers in our case. However, the algorithms proposed in the previous works do not apply in our setting. In our case, anticipating maneuvers requires modeling the interaction between the driving context and the driver's intention. Such interactions are absent in the previous works. We propose AIO-HMM to model these aspects of the problem.

\noindent \textbf{Computer vision for analyzing the human face.} 
The vision approaches related to our work are face detection and tracking~\citep{Viola04,Zhang10}, statistical models of face~\citep{Cootes01} and pose estimation methods for face~\citep{Xiong14}. Active Appearance Model (AAM)~\citep{Cootes01} and its variants~\citep{Matthews04,Xiong13} statistically model the shape and texture of the face. AAMs have also been used to estimate the 3D-pose of a face from a single image~\citep{Xiong14} and in design of assistive features for driver monitoring~\citep{Rezaei14,Tawari14b}. In our approach we adapt off-the-shelf available face detection and tracking algorithms for robustness required for anticipation (Section~\ref{sec:features}).
         
\noindent \textbf{Learning temporal models.} Temporal models are commonly used to model human activities~\citep{Koppula13c,Morency07,Wang06,Wang05}. These models have been used in both discriminative and generative fashions. The discriminative temporal models are mostly inspired by the Conditional Random Field (CRF)~\citep{Lafferty01} which  captures the temporal structure of the problem. Wang et al.~\citep{Wang05} and Morency et al.~\citep{Morency07} propose dynamic extensions of the CRF for image segmentation and gesture recognition respectively. The generative approaches for temporal modeling include various filtering methods, such as Kalman and particle filters~\citep{Thrun05}, Hidden Markov Models, and many types of Dynamic Bayesian Networks~\citep{Murphy12}. Some previous works~\citep{Berndt08,Kuge00,Oliver00} used HMMs to model different aspects of the driver's behaviour. Most of these generative approaches model how latent (hidden) states influence the observations. However, in our problem both the latent states and the observations influence each other. In particular, our AIO-HMM model is inspired by the Input-Output
HMM~\citep{Bengio95}. In the following sections we will explain the advantages of AIO-HMM over HMMs for anticipating maneuvers and also compare its performance with variants of HMM in the experiments
(Section~\ref{sec:experiment}).

%% file: overview.tex

\begin{figure*}
\centering
\vskip -0.17in
\includegraphics[width=.9\linewidth]{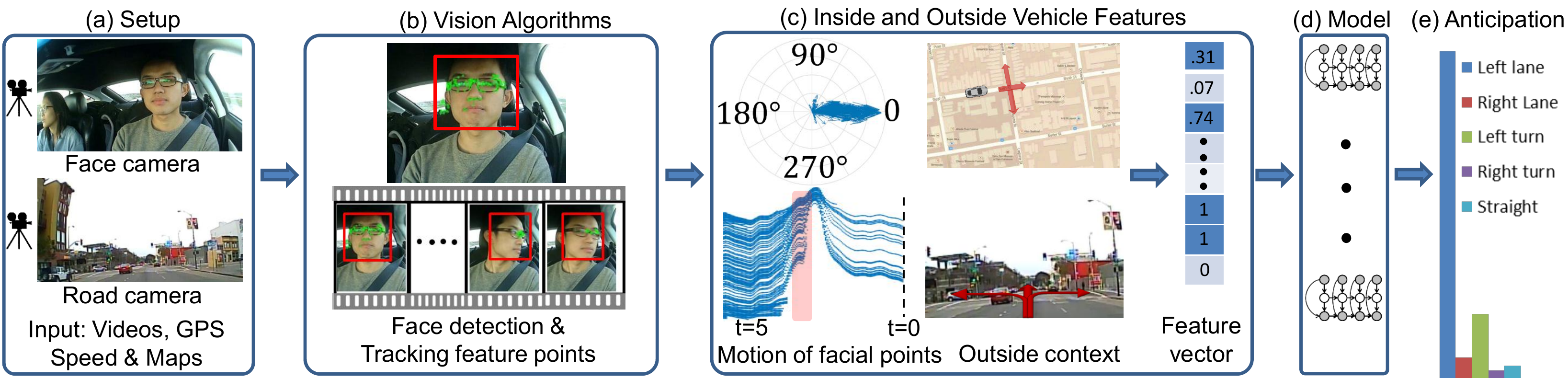}
\vspace{1.5\captionReduceTop}
\caption{\textbf{System Overview.} Our system anticipating a left lane
change maneuver. (a) We process multi-modal data including 
GPS, speed, street maps, and events inside and
outside of the vehicle using video cameras. (b) Vision pipeline extracts visual cues such as driver's head movements.
 (c) The inside and outside driving context is processed to extract expressive
 features. (d,e) Using our trained models we anticipate the probability of each maneuver.}
 \vspace{1.8\captionReduceBot}
\label{fig:system}
\end{figure*}

\vspace{1.4\sectionReduceTop}
\section{Problem Overview}
\vspace{\sectionReduceBot}
\label{sec:overview}
Our goal is to anticipate  driving maneuvers a few seconds before they occur. This includes anticipating a lane change before the wheels touch the lane markings or anticipating if the driver keeps straight or makes a turn when approaching an intersection. This is a challenging problem for multiple reasons. First, it requires the modeling of context from different sources. Information from a single source, such as a camera  capturing events outside the car, is not sufficiently rich.  Additional visual information from within the car can also be used.  For example, the driver's head movements are useful for anticipation -- drivers typically check for the side traffic while changing lanes and scan the cross traffic at intersections.

Second, reasoning about maneuvers should take into account the driving context at both local and global levels. Local context requires modeling events in vehicle's vicinity such as the surrounding vision, GPS, and speed information. On the other hand, factors that influence the overall route contributes to the global context, such as the driver's final destination. Third, the informative cues necessary for anticipation appear at variable times before the maneuver, as illustrated in Figure~\ref{fig:time_variance}.In particular, the time interval between the driver's head movement and the occurrence of the maneuver depends on factors such as the speed, traffic conditions, the GPS location, etc.

We obtain the driving context from different sources as shown in Figure~\ref{fig:system}. Our system includes: (1) a driver-facing camera inside the vehicle, (2) a road-facing camera outside the vehicle, (3) a speed logger, and (4) a GPS and map 
logger. The information from these sources constitute the \textit{driving context}.  We use the face camera to track the driver's head movements. The video from the road camera enables additional reasoning on maneuvers. For example, when the vehicle is in the left-most lane, the only safe maneuvers are a right-lane change or keeping straight, unless the vehicle is approaching an intersection. Maneuvers also correlate with the vehicle's speed, e.g., turns usually happen at  lower speeds than lane changes. Additionally, the GPS data augmented with the street map enables us to detect upcoming road artifacts such as intersections, highway exits, etc. We now describe our model and the learning algorithm.

%% file: model.tex
\vspace{1.75\sectionReduceTop}
\section{Our Approach}
\label{sec:approach}
\vspace{\sectionReduceBot}

\begin{figure}
\centering
\vskip -0.1in
\includegraphics[width=.85\linewidth]{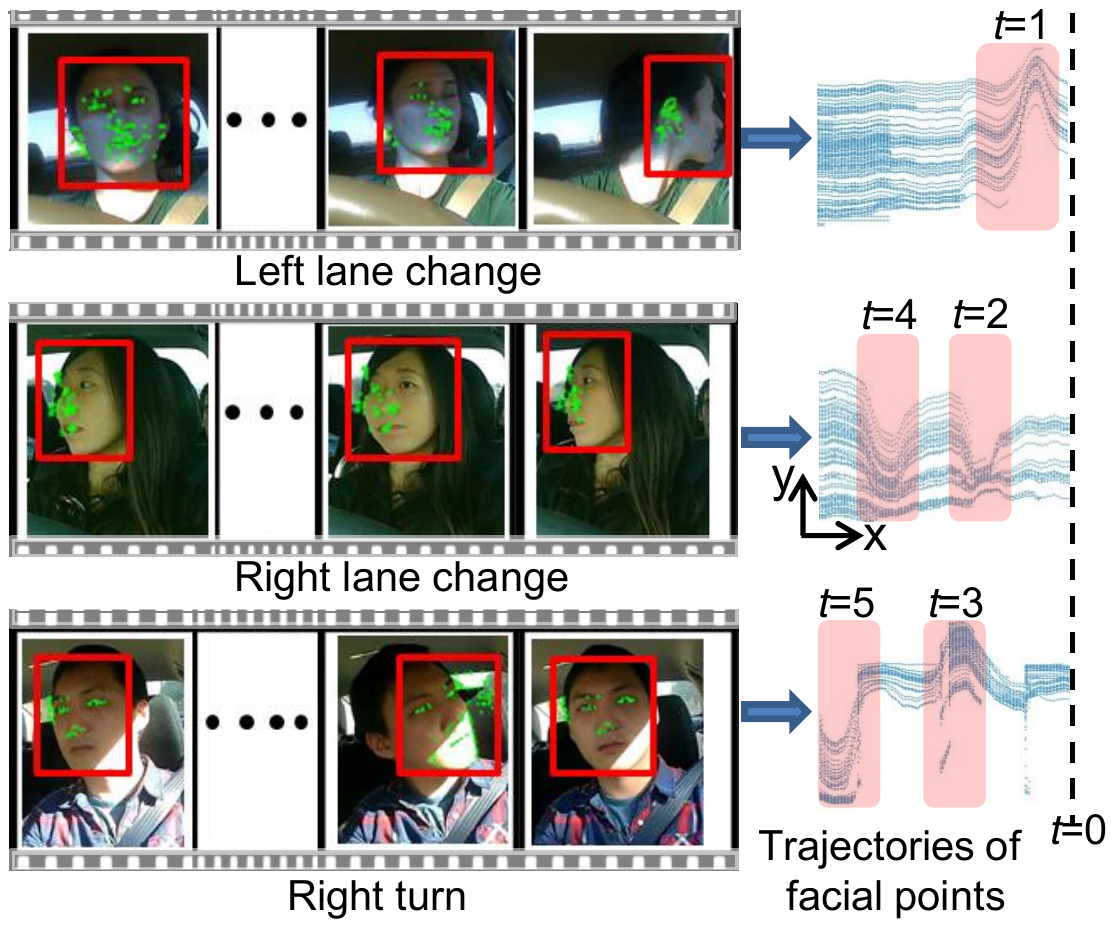}
\vspace{1.5\captionReduceTop}
\caption{\textbf{Variable time occurrence of events.} \textit{Left}: The events inside the vehicle before the maneuvers. We track the driver's face along with many facial points. \textit{Right}: The trajectories generated by the horizontal motion of facial points (pixels) `t' seconds before the maneuver. X-axis is the time and Y-axis is the pixels' horizontal coordinates. Informative cues appear during the shaded time interval. Such cues occur at variable times before the maneuver, and the order in which the cues appear is also important.}
\vspace{2\captionReduceBot}
\label{fig:time_variance}
\end{figure}
Driving maneuvers are influenced by multiple interactions involving the vehicle, its driver, outside traffic, and occasionally global factors like the driver's destination. These interactions influence the driver's intention, i.e. their state of mind before the maneuver, which is not directly observable.  We represent the driver's intention with discrete
states that are  \textit{latent} (or hidden). In order to anticipate maneuvers, we  jointly model the driving context and the  \textit{latent} states in a tractable manner. We represent the driving context as a set of features, which we describe in Section~\ref{sec:features}. We now present the motivation for our model and then describe the model, along with the learning and inference algorithms.

\vspace{1.5\subsectionReduceTop}
\subsection{Modeling driving maneuvers}
\vspace{1.5\subsectionReduceBot}
Modeling maneuvers require temporal modeling of the driving context (Figure~\ref{fig:time_variance}). Discriminative methods, such as the Support Vector Machine and the Relevance Vector Machine~\citep{Tipping01}, which do not model the temporal aspect perform poorly (shown in Section~\ref{subsec:results}). Therefore, a temporal model such as the Hidden Markov Model (HMM) is better suited. 

An HMM models how the driver's \textit{latent} states generate both the inside driving context and the outside driving context. However, a more accurate model should capture how events \textit{outside} the vehicle (i.e. the outside driving context) affect the driver's state of mind, which then generates the observations \textit{inside} the vehicle (i.e. the inside driving context). 
Such interactions can be modeled by an Input-Output HMM (IOHMM)~\citep{Bengio95}. However, modeling the problem with IOHMM will not capture the temporal dependencies of the inside driving context. These dependencies are critical to capture the smooth and temporally correlated behaviours such as the driver's face movements. We therefore present Autoregressive Input-Output HMM (AIO-HMM) which extends IOHMM to model these observation dependencies. Figure~\ref{fig:model} shows the AIO-HMM graphical model. 

\input{learning}

%% file: learning.tex
\vspace{1.5\subsectionReduceTop}
\subsection{Modeling Maneuvers with AIO-HMM}
\vspace{1.5\subsectionReduceBot}
\begin{figure}[t]
\centering
\vskip -0.075in
\includegraphics[width=.9\linewidth]{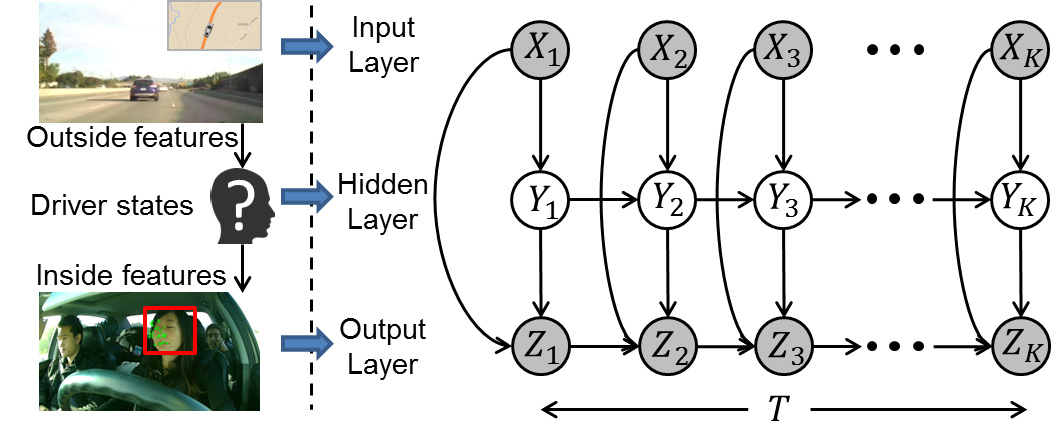}
\vspace{2\captionReduceTop}
\caption{\textbf{AIO-HMM.} The model has three layers: (i) Input (top): this  layer represents outside vehicle features $X$; (ii) Hidden (middle): this  layer represents driver's latent states $Y$; and (iii) Output (bottom): this  layer represents inside vehicle features $Z$. This layer also captures temporal dependencies of inside vehicle features. $T$ represents time.}
\vspace{2\captionReduceBot}
\label{fig:model}
\end{figure}

Given $T$ seconds long driving context $\mathcal{C}$  before the maneuver $M$, we learn a generative model for the context $P(\mathcal{C}|M)$. The driving context $\mathcal{C}$ consists of the outside driving context and the inside driving context. The outside and inside contexts are temporal sequences represented by the outside features $X_1^K = \{X_1,..,X_K\}$ and the inside features $Z_1^K = \{Z_1,..,Z_K\}$ respectively. The corresponding sequence of the driver's latent states is $Y_1^K = \{Y_1,..,Y_K\}$. $X$ and $Z$ are vectors and $Y$ is a discrete state.
\begin{align}
\nonumber P(\mathcal{C}|M) &= \sum_{Y_1^K} P(Z_1^K,X_1^K,Y_1^K|M)\\
\nonumber &= P(X_1^K|M)\sum_{Y_1^K} P(Z_1^K,Y_1^K|X_1^K,M)\\
\label{eq:aio-hmm} & \propto \sum_{Y_1^K} P(Z_1^K,Y_1^K|X_1^K,M)
\end{align}
We model the correlations between $X$, $Y$ and $Z$ with an AIO-HMM as shown in Figure~\ref{fig:model}. The AIO-HMM models the distribution in equation~\eqref{eq:aio-hmm}. It does not assume any generative process for the outside features $P(X_1^K|M)$. It instead models them in a discriminative manner. 
The top (input) layer of the AIO-HMM consists of  outside features $X_1^K$. The outside features then affect the driver's latent states $Y_1^K$, represented by the middle (hidden) layer, which then generates the inside features $Z_1^K$ at the bottom (output) layer. The events inside the vehicle such as the driver's head movements are temporally correlated because they are generally smooth. The AIO-HMM handles these dependencies with autoregressive connections in the output layer.  

\noindent \textbf{Model Parameters.}
AIO-HMM has two types of parameters: (i) state transition parameters $\mathbf{w}$; and (ii) observation emission parameters ($\boldsymbol\mu$,$\mathbf{\Sigma}$). We use set $\mathcal{S}$ to denote the possible latent states of the driver. For each state $Y=i\in\mathcal{S}$, we parametrize transition probabilities of leaving the state with log-linear functions, and parametrize the output layer feature emissions with normal distributions. 
\begin{align*}
\text{Transition: }& P(Y_t = j | Y_{t-1} = i,
X_t;\mathbf{w}_{ij}) = \frac{e^{\mathbf{w}_{ij} \cdot X_t}}{\sum_{l \in \mathcal{S}}
e^{\mathbf{w}_{il} \cdot X_t}}\\
\text{Emission: }&P(Z_t | Y_t=i,X_t,Z_{t-1};\boldsymbol\mu_{it},\mathbf{\Sigma}_i) =
\mathcal{N}(Z_t|\boldsymbol\mu_{it},\mathbf{\Sigma}_i) 
\end{align*}

The inside (vehicle) features represented by the output layer are jointly influenced by all three layers. These interactions are modeled by the mean and variance of the normal distribution. We model the mean of the distribution using the outside and inside features from the vehicle as follows:
$$\boldsymbol\mu_{it} = (1 + \mathbf{a}_i \cdot X_t + \mathbf{b}_i \cdot Z_{t-1})
\boldsymbol\mu_i$$
In the equation above, $\mathbf{a}_i$ and $\mathbf{b}_i$ are parameters that we learn for every state $i \in \mathcal{S}$. Therefore, the parameters we learn for state $i \in \mathcal{S}$ are $\boldsymbol\theta_i = \{\boldsymbol\mu_i$, $\mathbf{a}_i$, $\mathbf{b}_i$, $\mathbf{\Sigma}_i$ and $\mathbf{w}_{ij} | j \in \mathcal{S}\}$, and the overall model parameters are $\boldsymbol\Theta = \{\boldsymbol\theta_i|i\in\mathcal{S}\}$.

\vspace{1.5\subsectionReduceTop}
\subsection{Learning AIO-HMM parameters}
\vspace{1.5\subsectionReduceBot}

The training data $\mathcal{D} = \{(X_{1,n}^{K_n},Z_{1,n}^{K_n})|n=1,..,N\}$ consists of $N$ instances of a maneuver $M$. The goal is to maximize the data log-likelihood.
\begin{equation}
\label{eq:likelihood}
l(\boldsymbol\Theta;\mathcal{D}) = \sum_{n=1}^N
\log P(Z_{1,n}^{K_n}|X_{1,n}^{K_n};\boldsymbol\Theta)
 \end{equation}
Directly optimizing  equation~\eqref{eq:likelihood}  is challenging because parameters $Y$ representing the driver's states are \textit{latent}. We therefore use the iterative EM procedure to learn the model parameters. In EM, instead of directly maximizing equation~\eqref{eq:likelihood}, we maximize its simpler lower bound. We estimate the lower bound in the E-step and then maximize that estimate in the M-step. These two steps are repeated iteratively. 

\noindent \textbf{E-step.} In the E-step we get the lower bound of equation~\eqref{eq:likelihood} by calculating the expected
value of the \textit{complete} data log-likelihood using the current estimate of the parameter $\hat{\boldsymbol\Theta}$. 
\begin{equation}
\label{eq:estep}
\text{E-step: }Q(\boldsymbol\Theta;\hat{\boldsymbol\Theta}) =
E[l_c(\boldsymbol\Theta;\mathcal{D}_c)|\hat{\boldsymbol\Theta},\mathcal{D}] 
\end{equation}
where $l_c(\boldsymbol\Theta;\mathcal{D}_c)$ is the  log-likelihood 
 of  the \textit{complete} data $\mathcal{D}_c$ defined as:  
\begin{align}
\mathcal{D}_c &= \{(X_{1,n}^{K_n},Z_{1,n}^{K_n},Y_{1,n}^{K_n})|n=1,..,N\} \label{eq:complete-data}\\
l_c(\boldsymbol\Theta;\mathcal{D}_c) &= \sum_{n=1}^N
\log P(Z_{1,n}^{K_n},Y_{1,n}^{K_n}|X_{1,n}^{K_n};\boldsymbol\Theta) \label{eq:complete-likelihood}
\end{align}

We should note that the occurrences of hidden variables $Y$ in $l_c(\boldsymbol\Theta;\mathcal{D}_c)$ are marginalized in equation~\eqref{eq:estep}, and hence $Y$ need not be known.
We efficiently estimate $Q(\boldsymbol\Theta;\hat{\boldsymbol\Theta})$ using the forward-backward algorithm~\citep{Murphy12}. 

\noindent \textbf{M-step.} In the M-step we maximize the expected value  of the complete data log-likelihood $Q(\boldsymbol\Theta;\hat{\boldsymbol\Theta})$ and update the model parameter as follows:
\begin{equation}
\label{eq:mstep}
\text{M-step: }\boldsymbol\Theta = \argmax_{\boldsymbol\Theta}
Q(\boldsymbol\Theta;\boldsymbol\hat{\boldsymbol\Theta})
\end{equation}

Solving  equation~\eqref{eq:mstep} requires us to optimize for the parameters 
$\boldsymbol\mu$, $\mathbf{a}$, $\mathbf{b}$, $\mathbf{\Sigma}$ and $\mathbf{w}$. We optimize all parameters expect $\mathbf{w}$ exactly by deriving their closed form update expressions. We optimized $\mathbf{w}$ using the gradient descent. Refer to the supplementary material for detailed E and M steps.\footnote{\url{http://www.brain4cars.com/ICCVsupp.pdf}} 

\vspace{1.5\subsectionReduceTop}
\subsection{Inference of Maneuvers}
\vspace{1.5\subsectionReduceBot}
Our learning algorithm trains separate AIO-HMM models for each maneuver. The goal during inference is to determine which model best explains the past $T$ seconds of the driving context not seen during training. We evaluate the likelihood of the inside and outside feature sequences ($Z_1^K$ and $X_1^K$)  for each maneuver, and anticipate the probability $P_M$ of each maneuver $M$ as follows:
\begin{align}
\label{eq:infer} P_M &= P(M|Z_1^K, X_1^K) \propto P(Z_1^K, X_1^K | M)P(M)
\end{align}
Algorithm~\ref{alg:coactive} shows the complete inference procedure. The inference in equation~\eqref{eq:infer} simply requires a forward-pass~\citep{Murphy12} of the AIO-HMM, the complexity of which is 
$\mathcal{O}(K(|\mathcal{S}|^2 + |\mathcal{S}||Z|^3 + |\mathcal{S}||X|))$. However, in practice it is only $\mathcal{O}(K|\mathcal{S}||Z|^3)$ because $|Z|^3 \gg |S|$ and $|Z|^3 \gg |X|$. Here $|\mathcal{S}|$
is the number of discrete states representing the driver's intention, while $|Z|$ and $|X|$ are the dimensions of the inside and outside feature vectors respectively. In equation~\eqref{eq:infer} $P(M)$ is the prior probability of maneuver $M$. We assume an uninformative uniform prior over the maneuvers.
  \begin{algorithm}[H]\caption{Anticipating maneuvers }
\begin{algorithmic}
\INPUT Driving videos, GPS, Maps and Vehicle Dynamics
\OUTPUT Probability of each maneuver
\STATE Initialize the face tracker with the driver's face
\WHILE{$driving$}
	\STATE Track the driver's face~\citep{Viola04}
	\STATE Extract features $Z_1^K$ and $X_1^K$ (Sec.~\ref{sec:features})
	\STATE Inference $P_M = P(M|Z_1^K, X_1^K)$ (Eq.~\eqref{eq:infer})
	\STATE Send the inferred probability of each maneuver to ADAS
\ENDWHILE
\end{algorithmic}
\label{alg:coactive}
\end{algorithm}

%% file: features.tex
\vspace{3.5\sectionReduceTop}
\section{Features}
\label{sec:features}
\vspace{\sectionReduceBot}

\begin{figure}[t]
\centering
\includegraphics[width=.9\linewidth]{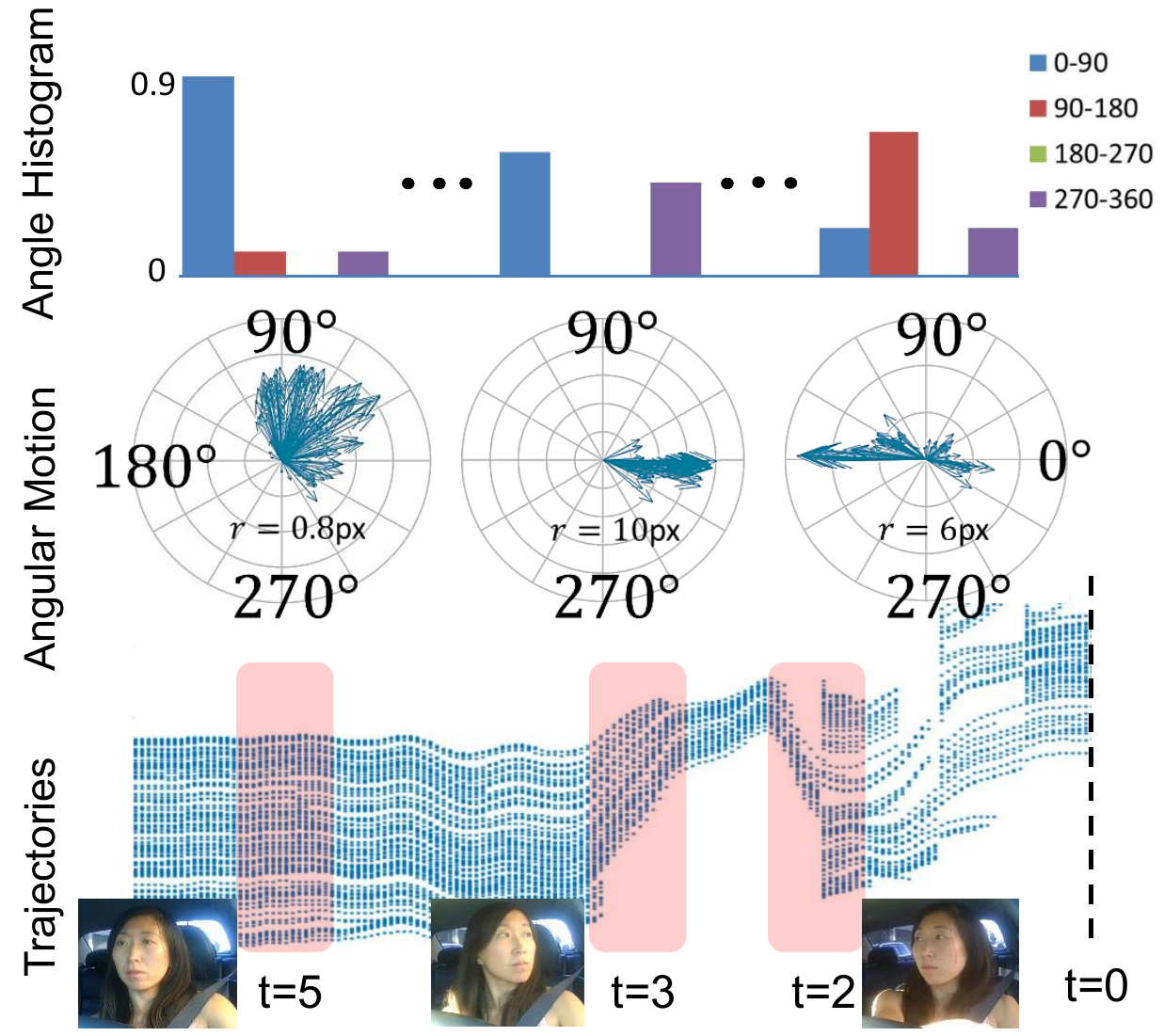}
\vspace{0.1\sectionReduceTop}
\caption{\textbf{Inside vehicle feature extraction.} The angular histogram features extracted at three different time steps for a left turn maneuver. \textit{Bottom}: Trajectories for the horizontal motion of tracked facial pixels `t' seconds before the maneuver. At t=5 seconds before the maneuver the driver is looking straight, at t=3 looks (left) in the direction of maneuver, and at t=2 looks (right) in opposite direction for the crossing traffic. \textit{Middle}: Average motion vector of tracked facial pixels in polar coordinates. $r$ is the average movement of pixels and arrow indicates the direction in which the face moves when looking from the camera. \textit{Top}: Normalized angular histogram features. }
\vspace{3\sectionReduceBot}
\label{fig:face_feature}
\end{figure}
We extract features by processing the inside and outside driving contexts. We denote the inside features with $Z$ and the outside features with $X$.
\vspace{1\subsectionReduceTop}
\subsection{Inside-vehicle features.}\label{subsec:inside_features} 
\vspace{1.5\subsectionReduceBot}
The inside features $Z$ capture the driver's head movements. Our vision pipeline consists of face detection, tracking, and feature extraction modules. We extract head motion features per-frame,  denoted by  $\phi(\text{face})$. For AIO-HMM, we compute $Z$ by aggregating  $\phi(\text{face})$ for every 20 frames, i.e., $Z = \sum_{i=1}^{20}\phi(\text{face}_i)/\|\sum_{i=1}^{20}\phi(\text{face}_i)\|$. 
\vspace{0.05in}

\noindent \textit{Face detection and tracking.} We detect
the driver's face using a trained Viola-Jones face detector~\citep{Viola04}. From the detected face, we first extract visually discriminative (facial) points using the Shi-Tomasi corner
detector \citep{Shi94} and then track those facial points using the Kanade-Lucas-Tomasi tracker \citep{Lucas81,Shi94,Tomasi91}. However, the tracking may accumulate errors over time because of changes in illumination due to the shadows of trees,  traffic, etc. We therefore constrain the tracked facial points to follow a projective transformation and remove the incorrectly tracked points using the RANSAC algorithm. While tracking the facial points, we  lose some of the tracked  points with every new frame. To address this problem, we  re-initialize the tracker with new discriminative facial points once the number of tracked  points falls below a threshold \citep{Kalal10}. 
\vspace{0.05in}

\noindent \textit{Head motion features.} 
For maneuver anticipation the horizontal movement of the face and its angular rotation (\textit{yaw}) are particularly important.  From the face tracking we obtain \textit{face tracks}, which are 2D trajectories of the tracked facial points in the image plane. Figure~\ref{fig:face_feature} (bottom) shows how the horizontal coordinates of the tracked facial points vary with time before a left turn maneuver. We represent the driver's face movements and rotations with histogram features. In particular, we take matching facial points between  successive frames and create histograms of their corresponding horizontal motions (in pixels) and angular motions in the image plane (Figure~\ref{fig:face_feature}). We bin the horizontal and angular motions using $[\leq-2,\;-2\;\text{to}\;0,\;0\;\text{to}\;2,\;\geq2]$ and  $[0\;\text{to}\;\frac{\pi}{2},\;\frac{\pi}{2}\;\text{to}\;\pi,\;\pi\;\text{to}\;\frac{3\pi}{2},\;\frac{3\pi}{2}\;\text{to}\;2\pi]$,
respectively. We also calculate the mean movement of the driver's face center. This gives us $\phi(\text{face})\in\mathbb{R}^9$ facial features per-frame. The driver's eye-gaze is also useful a feature. However, robustly estimating 3D eye-gaze in outside environment is still a topic of research, and orthogonal to this work on anticipation. We therefore do not consider eye-gaze features.  

\vspace{1.5\subsectionReduceTop}
\subsection{Outside-vehicle features.}\label{subsec:outside_features} 
\vspace{1.5\subsectionReduceBot}
The outside feature vector $X$ encodes the information about the outside environment such as the road conditions, vehicle dynamics, etc. In order to get this information, we use the road-facing camera together with the vehicle's GPS coordinates, its speed, and the street maps. More specifically, we obtain two binary features from the road-facing camera   indicating whether a lane exists on the left side and on the right side of the  vehicle. We also augment the vehicle's GPS coordinates with the street maps and extract a binary feature indicating if the vehicle is within 15 meters of a road artifact  such as intersections, turns, highway exists, etc. We also encode the average,  maximum, and  minimum speeds of the vehicle over the last 5 seconds as features. This results in a $X \in\mathbb{R}^6$ dimensional feature vector. 

%% file: experiment.tex

\vspace{\sectionReduceTop}
\section{Experiment}
\label{sec:experiment}
\vspace{\sectionReduceBot}

\begin{figure}
\vskip -0.1in
\centering
\includegraphics[width=.9\linewidth]{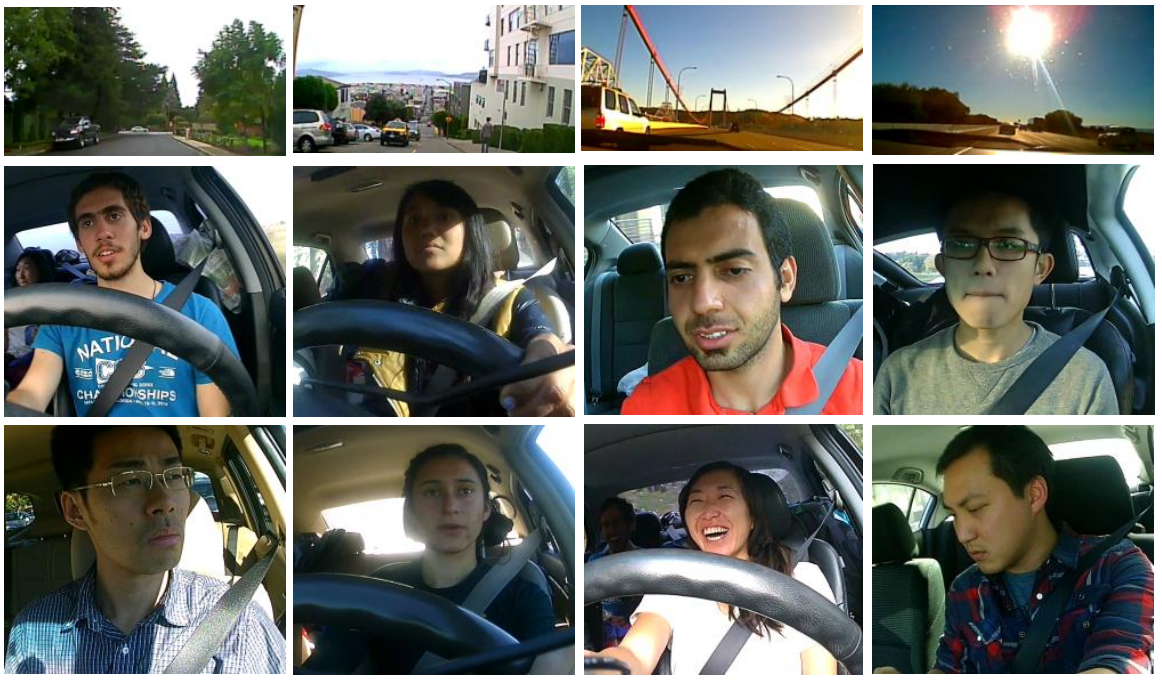}
\vspace{1\captionReduceTop}
\caption{\textbf{Our data set} is diverse in drivers and landscape.}
\vspace{2\captionReduceBot}
\label{fig:diverse_data}
\end{figure}

\begin{table*}[t]
\centering
\vskip -0.1in
\caption{\textbf{Results on our driving data set}, showing average \textit{precision}, \textit{recall} and \textit{time-to-maneuver} computed from 5-fold cross-validation. The number inside parenthesis is the standard error.}
\vskip -0.1in
\resizebox{.8\textwidth}{!}{
\centering
\begin{tabular}{r|ccc|ccc|ccc}\hline
  &\multicolumn{3}{c}{Lane change}&\multicolumn{3}{|c}{Turns}&\multicolumn{3}{|c}{All maneuvers}\\
\cline{2-10}
& \multirow{2}{*}{$Pr$ (\%)}  & \multirow{2}{*}{$Re$ (\%)} & Time-to-  & \multirow{2}{*}{$Pr$ (\%)} & \multirow{2}{*}{$Re$ (\%)} & Time-to-  & \multirow{2}{*}{$Pr$ (\%)} & \multirow{2}{*}{$Re$ (\%)}  & Time-to- \\ 
Algorithm & & &  maneuver (s) &  & &  maneuver (s) &  & & maneuver (s)\\\hline
Chance	&	33.3		&	33.3		&	-	&	33.3		&	33.3		&	-	&	20.0		&	20.0		&	-\\
Morris et al.~\citep{Morris11} SVM	&	73.7 (3.4)	&	57.8 (2.8)	&	2.40 (0.00)		&	64.7 (6.5)	&	47.2 (7.6)	&	2.40 (0.00)		&	43.7 (2.4)	&	37.7 (1.8)	& 1.20 (0.00)\\
Random-Forest &   71.2 (2.4)  &   53.4 (3.2)  &   3.00 (0.00)   &   68.6 (3.5)  &
44.4 (3.5)  &   1.20 (0.00)      &   51.9 (1.6)  &   27.7 (1.1)  & 1.20 (0.00)\\
HMM $E$ & 75.0 (2.2) & 60.4 (5.7) & 3.46 (0.08) &  74.4 (0.5) & 66.6 (3.0) & 4.04 (0.05) & 63.9 (2.6) & 60.2 (4.2) & 3.26 (0.01)\\
HMM $F$		&	76.4 (1.4)	&	75.2 (1.6)	&	3.62 (0.08)		&	75.6 (2.7)	&		60.1 (1.7)	&	3.58 (0.20)		&	64.2 (1.5)	&	36.8 (1.3)	&	2.61 (0.11)\\
HMM $E+F$	&	80.9 (0.9)	&	79.6 (1.3)	&	3.61 (0.07)		&	73.5 (2.2)	&
75.3 (3.1)	&	4.53 (0.12)		&	67.8 (2.0)	&	67.7 (2.5)	&	3.72 (0.06)\\\hline
\textit{(Our method)} IOHMM		&	81.6 (1.0)	&	\textbf{79.6 (1.9)}	&	3.98  {(0.08)}	&	77.6 (3.3)	&		\textbf{75.9 (2.5)}	&	4.42 {(0.10)}		&	74.2 (1.7)	&	71.2 (1.6)	&	3.83 {(0.07)}\\
\textit{(Our final method)} AIO-HMM		&	\textbf{83.8 (1.3)}	&	79.2 (2.9)	&	3.80  {(0.07)}		&	\textbf{80.8	 (3.4)}	&		75.2 (2.4)	&	4.16 {(0.11)}		&	\textbf{77.4 (2.3)}	&	\textbf{71.2 (1.3)}	&	3.53 {(0.06)}\\\hline
\end{tabular}
}
\label{tab:prscore}
\end{table*}
\begin{figure*}[t]
\vskip -0.08in
\centering
\begin{subfigure}[b]{.22\textwidth}
	\includegraphics[width=0.8\linewidth]{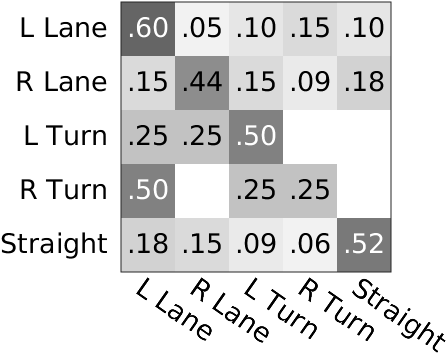}
	\vspace{\captionReduceTop}
	\caption{SVM~\citep{Morris11}}
	\vspace{.3\captionReduceTop}
	\end{subfigure}
\begin{subfigure}[b]{.22\textwidth}
	\includegraphics[width=.8\linewidth]{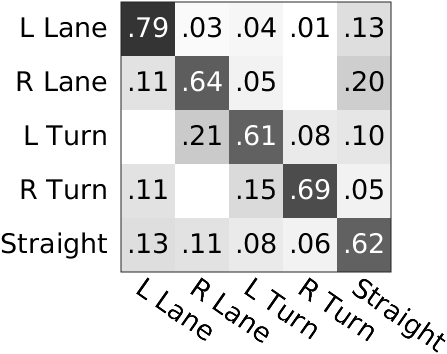}
	\vspace{\captionReduceTop}
	\caption{HMM $E+F$}
	\vspace{.3\captionReduceTop}
	\end{subfigure}	
	\begin{subfigure}[b]{.22\textwidth}
	\includegraphics[width=0.8\linewidth]{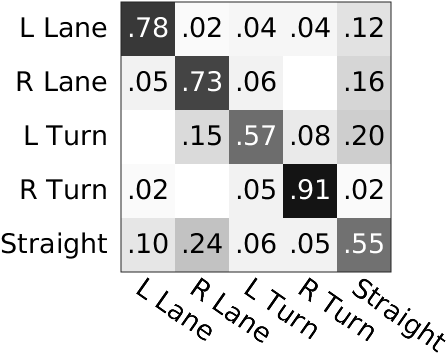}
	\vspace{\captionReduceTop}
	\caption{IOHMM}
	\vspace{.3\captionReduceTop}
	\end{subfigure}	
	\begin{subfigure}[b]{.22\textwidth}
	\includegraphics[width=0.8\linewidth]{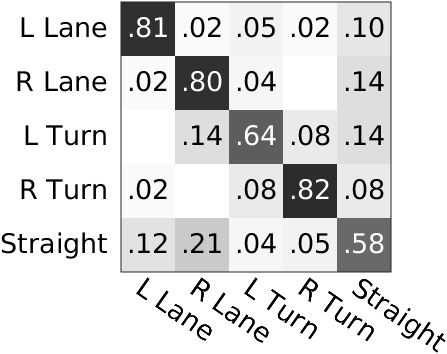}
	\vspace{\captionReduceTop}
	\caption{AIO-HMM}
	\vspace{.3\captionReduceTop}
	\end{subfigure}
	\vskip -0.1in
	\caption{\textbf{Confusion matrix} of different algorithms when jointly predicting all the maneuvers. Predictions made by algorithms are represented by rows, actual maneuvers are represented by columns, and precision on diagonal.}
	\vspace{2\captionReduceBot}
	\label{fig:confmat}	
\end{figure*}
We first give an overview of our data set, the baseline algorithms, and our evaluation setup. We then present the results and discussion. Our video demonstration is available at: \url{http://www.brain4cars.com}. 

\vspace{1.5\subsectionReduceTop}
\subsection{Experimental Setup}
\vspace{2\subsectionReduceBot}
\noindent \textbf{Data set.}
Our data set consists of natural driving videos with both inside and outside views of the car, its speed, and the global position system (GPS) coordinates.\footnote{The inside and outside cameras operate at 25 and 30 frames/sec.} The outside car video captures the view of the road ahead. We collected this driving data set under fully natural settings without any intervention.\footnote{\textbf{Protocol:} We set up cameras, GPS and speed recording device in subject's personal vehicles and left it to record the data. The subjects were asked to ignore our setup and drive as they would normally.} 
It consists of 1180 miles of freeway and city driving and encloses 21,000 square miles across two states. We collected this data set from 10 drivers over a period of two months. The complete data set has a total of 2 million video frames and includes diverse landscapes. Figure~\ref{fig:diverse_data} shows a few samples from our data set. We annotated the driving videos with a total of 700 events containing 274 lane changes, 131 turns, and 295 randomly sampled instances of driving straight. Each lane change or turn annotation marks the start time of the maneuver, i.e., before the car touches the lane or yaws, respectively. For all annotated events, we also annotated the lane information, i.e., the number of lanes on the road and the current lane of the car.

\noindent\textbf{Baseline algorithms} we compare with: 
\vskip -0.01in
\begin{itemize}
\item \textit{Chance:} Uniformly randomly anticipates a maneuver.
\item \textit{SVM~\citep{Morris11}:} Support Vector Machine is a discriminative classifier~\citep{Cortes95}. Morris et al.~\citep{Morris11} takes this approach  for anticipating maneuvers.\footnote{Morries et al.~\citep{Morris11} considered binary 
classification problem (lane change vs driving straight) and used RVM~\citep{Tipping01}.}
We train the SVM on 5 seconds of driving context by concatenating all frame features 
to get a $\mathbb{R}^{3840}$ dimensional feature vector. 
\item \textit{Random-Forest~\citep{Criminisi11}:} This is also a discriminative classifier that
learns many decision trees from the training data, and at test time it
averages the prediction of the individual decision trees. We train
it on the same features as SVM with 150 trees of depth ten each.
\item \textit{HMM:} This is the Hidden Markov Model. 
We train the HMM on a temporal sequence of feature vectors that we extract every 0.8 seconds, 
i.e., every 20 video frames.  
We consider three versions of the HMM: (i) HMM $E$: with only outside features from the road camera, the vehicle's speed, GPS and street maps (Section~\ref{subsec:outside_features});
(ii) HMM $F$: with only inside features from the driver's face (Section~\ref{subsec:inside_features}); 
and (ii) HMM $E+F$: with both inside and outside features. 
\end{itemize}
We compare these baseline algorithms with our IOHMM and AIO-HMM models. The features for our model are extracted  in the same manner as in HMM $E+F$ method.

\noindent \textbf{Evaluation setup.} We evaluate an algorithm based on its correctness in predicting future maneuvers.
 We anticipate maneuvers every 0.8 seconds where the algorithm 
processes the recent context and assigns a probability to each of the four maneuvers: \{\textit{left lane 
change, right lane change, left turn, right turn}\} and a probability to the event of \textit{driving straight}. 
 These five probabilities together sum to one.
After anticipation, i.e. when the algorithm has computed all five probabilities, the algorithm predicts a  maneuver if its probability is above a threshold. If none of the maneuvers' probabilities are above this threshold, the algorithm does not make a maneuver prediction and predicts \textit{driving  straight}.
 However, when it predicts 
one of the four maneuvers, it sticks with this prediction and makes no further predictions for next 5 
seconds or until a maneuver occurs, whichever happens earlier. After 5 seconds or a 
maneuver has occurred, it returns to anticipating  future maneuvers. 

During this process of anticipation and prediction, the algorithm makes (i) true predictions ($tp$): when it predicts the correct maneuver; (ii) false predictions ($fp$): when it predicts a maneuver but the driver performs a different  maneuver; (iii) false positive predictions ($fpp$): when it predicts a maneuver but the driver does not perform any maneuver (i.e. \textit{driving straight}); and (iv) missed predictions ($mp$): when it predicts \textit{driving straight} but the driver performs a maneuver. We evaluate the algorithms using their precision and recall scores:
$$Pr = \frac{tp}{\underbrace{tp+fp+fpp}_\text{Total \# of maneuver predictions}};\;\;\;Re=\frac{tp}{\underbrace{tp+fp+mp}_\text{Total \# of maneuvers}}$$
The precision measures the fraction of the predicted maneuvers that are correct and recall measures the 
fraction of the maneuvers that are correctly predicted. For true predictions ($tp$) we also compute the 
average \textit{time-to-maneuver}, where time-to-maneuver is  the interval between the  time of algorithm's prediction and the start of the maneuver.

We perform cross validation to choose the number of the driver's latent states in the AIO-HMM and the threshold on  probabilities for maneuver prediction. For \textit{SVM} we cross-validate for the parameter $C$ and the choice of kernel from Gaussian and polynomial kernels. The parameters are chosen as the ones giving the highest F1-score on a validation set. 
The F1-score is the harmonic mean of the precision and recall, defined as $F1 = 2*Pr*Re/(Pr+Re)$.

\vspace{1.8\subsectionReduceTop}
\subsection{Results and Discussion}
\label{subsec:results}
\vspace{2\subsectionReduceBot}
We evaluate the algorithms on maneuvers that were not seen during training  and report the results using 5-fold cross validation. Table \ref{tab:prscore} reports the precision and recall scores 
under three settings: (i) \textit{Lane change}: when the algorithms only predict for the left and right lane changes. This setting is relevant for highway driving where the prior probabilities of turns are low; (ii) \textit{Turns}: when the algorithms only predict for the left and right turns; and (iii) \textit{All maneuvers}: here the algorithms jointly predict all four maneuvers. All three settings include the instances of \textit{driving straight}. 

As shown in Table~\ref{tab:prscore}, the AIO-HMM performs better than the other algorithms. 
Its precision is over 80\% for the \textit{lane change} and \textit{turns} settings. 
For jointly predicting all the maneuvers its precision is 77\%, which is 34\% higher than the 
previous work by Morris et al.~\citep{Morris11} and 26\% higher than the Random-Forest. The AIO-HMM recall is always comparable or 
better than the other algorithms.
On average the AIO-HMM predicts maneuvers 3.5 seconds before they occur and up to 4 seconds earlier when only predicting turns.

Figure~\ref{fig:confmat} shows the confusion matrix plots for jointly anticipating all the maneuvers. AIO-HMM gives the highest precision for each maneuver. Modeling maneuver anticipation with an input-output model allows for a discriminative modeling of the state transition probabilities using rich features from outside the vehicle. On the other hand, the HMM $E+F$ solves a harder problem by learning a generative model of the outside and inside features together. As shown in Table~\ref{tab:prscore}, the precision of HMM $E+F$ is 10\% less than that of AIO-HMM for jointly predicting all the maneuvers. AIO-HMM extends IOHMM by modeling the temporal dependencies of events inside the vehicle. This results in better performance: on average AIO-HMM precision is 3\% higher than IOHMM, as shown in Table~\ref{tab:prscore}.
 
\begin{figure}[t]
\vskip -0.15in
\centering
\includegraphics[width=.65\linewidth]{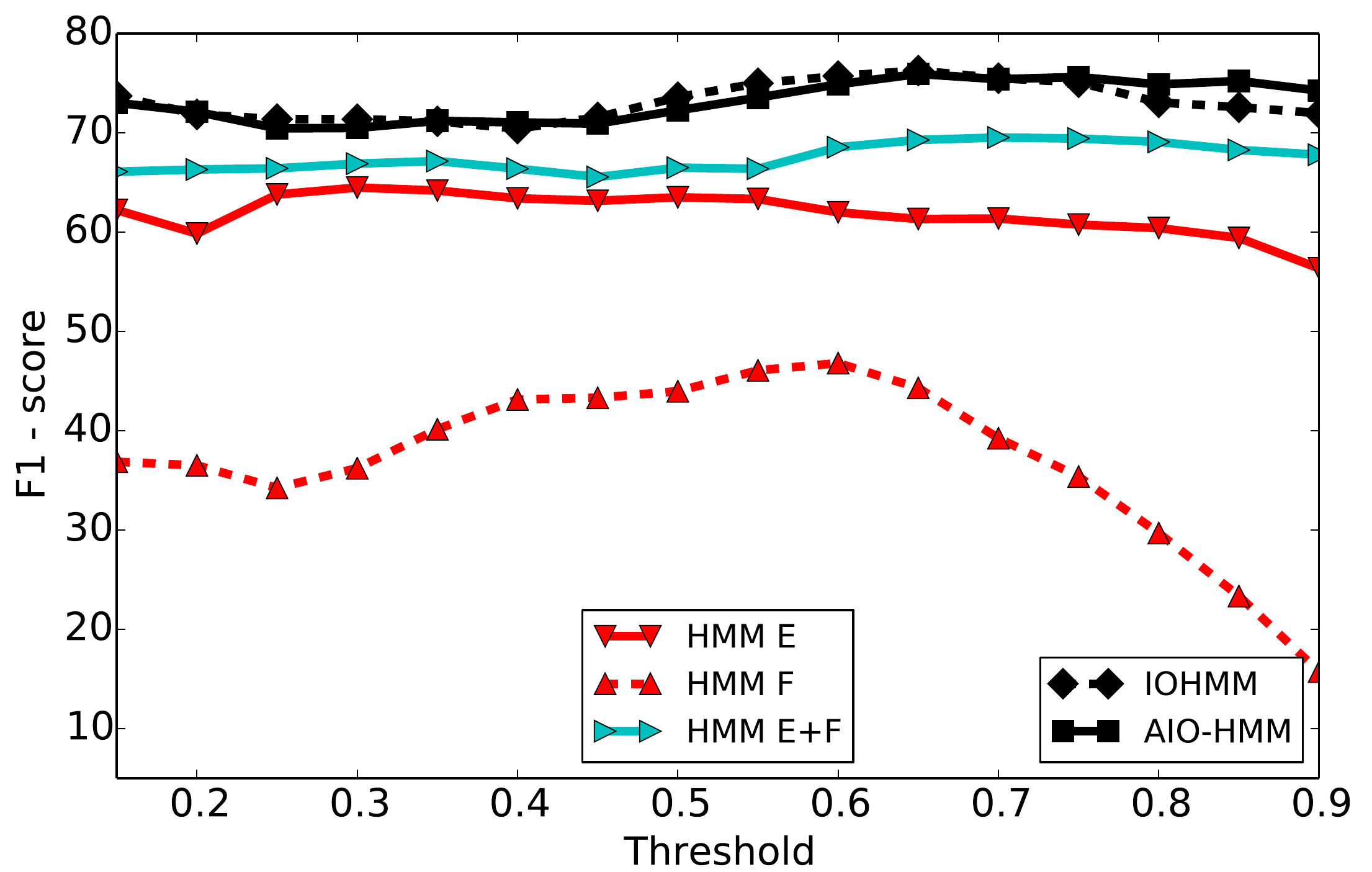}
\vspace{1\captionReduceBot}
\caption{\textbf{Effect of prediction threshold.} An
algorithm makes a prediction only when its confidence is above the prediction 
threshold. Plot shows how the F1-score varies with prediction threshold.}
\vspace{2\captionReduceBot}
\label{fig:threshold}
\end{figure}

Table~\ref{tab:fpp} compares the $fpp$ of different algorithms.  False positive predictions ($fpp$) happen when an algorithm wrongly predicts \textit{driving straight} as one of the maneuvers. Therefore low value of $fpp$ is preferred. HMM $F$ performs best on this metric at 11\%
as it mostly assigns a high probability to \textit{driving straight}. 
However, due to this reason, it incorrectly predicts \textit{driving straight} even when maneuvers happen. This results in the low recall of HMM $F$ at 36\%, as shown in Table~\ref{tab:prscore}. AIO-HMM's 
$fpp$ is 10\% less than that of IOHMM and HMM $E+F$. In Figure~\ref{fig:threshold} we compare the F1-scores of different algorithms as the prediction threshold varies. We observe that IOHMM and AIO-HMM perform better than the baseline algorithms and their F1-scores remains stable as we vary the threshold. Therefore, the prediction threshold is useful as a parameter to trade-off between the precision and recall of algorithms.
 
\begin{table}[t]
\centering
\vskip -0.15in
\caption{\textbf{False positive prediction} ($fpp$) of different algorithms. The number inside parenthesis is the standard error.}
\vspace{.8\captionReduceBot}
\resizebox{.85\linewidth}{!}{
\begin{tabular}{r|ccc}
Algorithm	&	Lane change	&	Turns	&	All\\\hline
Morris et al.~\citep{Morris11} SVM	&	15.3 (0.8) &		13.3 (5.6)	&	24.0 (3.5)\\
Random-Forest & 16.2 (3.3) & 12.9 (3.7) & 17.5 (4.0) \\
HMM $E$ & 36.2 (6.6) & 33.3 (0.0) & 63.8 (9.4) \\
HMM $F$	&	23.1 (2.1)	&	23.3 (3.1)	&	11.5 (0.1)\\
HMM $E+F$	&	30.0 (4.8)	&	21.2 (3.3)	&	40.7	 (4.9)\\
IOHMM	&	28.4 (1.5)	&	25.0 (0.1)	&	40.0 (1.5)\\
AIO-HMM	&	24.6 (1.5)	&	20.0 (2.0)	&	30.7 (3.4)\\
\end{tabular}
}
\vspace{0\captionReduceBot}
\label{tab:fpp}
\end{table}
\begin{figure}[t]
\vskip -0.2in
\centering
\includegraphics[width=.67\linewidth]{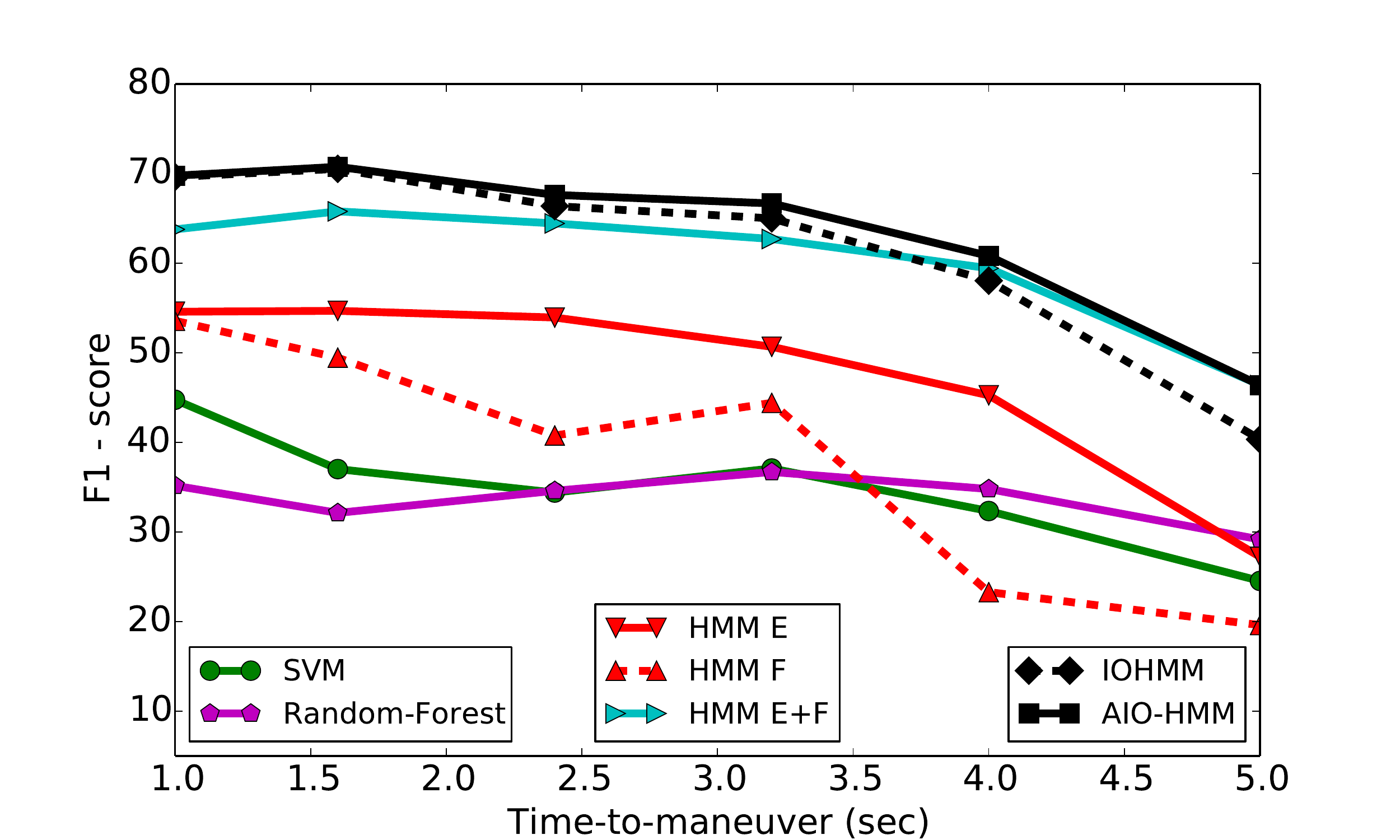}
\vspace{0.7\captionReduceBot}
\caption{\textbf{Effect of time-to-maneuver.} Plot compares $F1$-scores when algorithms predict maneuvers at a fixed time-to-maneuver, and shows change in performance as we vary time-to-maneuver.}
\vspace{2.3\captionReduceBot}
\label{fig:time_predict}
\end{figure}

\noindent\textbf{Importance of inside and outside driving context.} 
An important aspect of anticipation is the joint modeling of the inside and outside driving contexts. HMM $F$ models only the inside driving context, while HMM $E$ models only the outside driving context.
As shown in Table~\ref{tab:prscore}, the precision and recall values of both models is less than HMM $E+F$, which jointly models both the contexts. More specifically, the precision of HMM $F$ on jointly predicting all the maneuvers in 3\%, 10\%, and 13\% 
less than  that of HMM $E+F$, IOHMM, and AIO-HMM, respectively. For HMM $E$ this difference is 4\%, 11\%, and 14\% respectively.

\noindent\textbf{Effect of time-to-maneuver.}
In Figure~\ref{fig:time_predict} we compare F1-scores of the algorithms when they predict maneuvers at a fixed time-to-maneuver, and show how the performance changes as we vary the time-to-maneuver. As we get closer to the start of the maneuvers the F1-scores of the algorithms increase. As opposed to this setting, in Table~\ref{tab:prscore} the algorithms predicted maneuvers at the time they were most confident. Under  both the fixed and variable time prediction settings, the AIO-HMM performs better than the baselines.

\noindent\textbf{Anticipation complexity.} The AIO-HMM anticipates maneuvers every 0.8 seconds using the previous 5 seconds of the driving context. The complexity mainly comprises of feature extraction and the model inference in equation~\eqref{eq:infer}. Fortunately both these steps can be performed as a dynamic program by storing the computation of the most recent anticipation. Therefore, for every anticipation we only process the incoming 0.8 seconds and not complete 5 seconds of the driving context. Due to dynamic programming the inference complexity described in equation~\eqref{eq:infer}, $\mathcal{O}(K|\mathcal{S}||I|^3)$, no longer depends on $K$ and reduces to $\mathcal{O}(|\mathcal{S}||I|^3)$.  On average we predict a maneuver under 3.6~milliseconds on a 3.4GHz CPU using MATLAB~2014b on Ubuntu~12.04. 

\vspace{1.5\subsectionReduceTop}
\subsection{Qualitative discussion}
\label{subsec:discuss}
\vspace{2\subsectionReduceBot}
\noindent \textbf{Common Failure Modes.} Wrong anticipations can occur for different reasons. These include failures in the vision pipeline and unmodeled events such as interactions with fellow passengers, overtakes, etc. In 6\% of the maneuvers, our tracker failed due to changes in illumination (in supplementary we show some instances). Wrong anticipations are also common when drivers strongly rely upon their recent memory of traffic conditions. In such situations visual cues are partially available in form of eye movements. Similarly, when making turns from turn-only lanes drivers tend not to reveal many visual cues. With rich sensory integration, such as radar for modeling the traffic, infra-red cameras for eye-tracking, along with reasoning about the traffic rules, we can further improve the performance. 
Fortunately, the automobile industry has made 
significant advances in some of these areas~\citep{AudiSelfDriving,GoogleSelfDriving,Wang15}  where our work can apply. 
Future work also includes extending our approach to night driving.

\noindent \textbf{Prediction timing.} In anticipation there is an inherent ambiguity. Once the algorithm is certain about a maneuver above a threshold probability should it predict immediately or should it wait for more information? An example of this ambiguity is in situations where drivers scan the traffic but do not perform a maneuver. In such situations different prediction strategies will result in different performances.

%% file: conclusions.tex
\vspace{2.3\sectionReduceTop}
\section{Conclusion}
\vspace{1.5\sectionReduceBot}
In this paper we considered the problem of anticipating driving maneuvers a few seconds before the driver performs them. Our work enables advanced driver assistance systems (ADAS) to alert drivers before they perform a dangerous maneuver, thereby giving drivers more time to react.  
We proposed an AIO-HMM to  jointly model the driver's intention and the driving context from both inside and outside of the car. Our approach accurately handles both the temporal and generative nature of the problem. 
We extensively evaluated on 1180 miles of driving data and showed improvement over many baselines. Our inference takes only a few milliseconds  therefore  it is suited for real-time use.  We will also publicly release our data set of natural driving. 
\\~
\textbf{Acknowledgement.} This work was supported by NRI award 1426452, ONR award N00014-14-1-0156, and by Microsoft Faculty Fellowship and NSF Career Award to one of us (Saxena).